\documentclass[letterpaper]{article} % DO NOT CHANGE THIS
\pdfoutput=1
\usepackage{aaai24}  % DO NOT CHANGE THIS
\usepackage{times}  % DO NOT CHANGE THIS
\usepackage{helvet}  % DO NOT CHANGE THIS
\usepackage{courier}  % DO NOT CHANGE THIS
\usepackage[hyphens]{url}  % DO NOT CHANGE THIS
\usepackage{graphicx} % DO NOT CHANGE THIS
\usepackage{natbib}  % DO NOT CHANGE THIS AND DO NOT ADD ANY OPTIONS TO IT
\usepackage{caption} % DO NOT CHANGE THIS AND DO NOT ADD ANY OPTIONS TO IT
\usepackage{tikz}
\usepackage{amsmath}
\usepackage[ruled,linesnumbered]{algorithm2e}
\usepackage{multirow}
\usepackage{booktabs}
\usepackage{subfigure}
\nocopyright

\title{Parameter-Saving Adversarial Training: Reinforcing Multi-Perturbation Robustness via Hypernetworks}
\author {
    Huihui Gong\textsuperscript{\rm 1},
    Minjing Dong\textsuperscript{\rm 1},
    Siqi Ma\textsuperscript{\rm 2},
    Seyit Camtepe\textsuperscript{\rm 3},
    Surya Nepal\textsuperscript{\rm 3},
    Chang Xu\textsuperscript{\rm 1},
}
\affiliations {
    \textsuperscript{\rm 1}School of Computer Science, University of Sydney\\
    \textsuperscript{\rm 2}School of Engineering and Information Technology, University of New South Wales, Canberra\\
    \textsuperscript{\rm 3}CISRO Data61\\
    \{hgon9611,mdon0736\}@uni.sydney.edu.au, siqi.ma@adfa.edu.au, \{Seyit.Camtepe,surya.nepal\}@data61.csiro.au, c.xu@sydney.edu.au
}

\usepackage{bibentry}
\begin{document}

\maketitle

\begin{abstract}
Adversarial training serves as one of the most popular and effective methods to defend against adversarial perturbations. However, most defense mechanisms only consider a single type of perturbation while various attack methods might be adopted to perform stronger adversarial attacks against the deployed model in real-world scenarios, e.g., $\ell_2$ or $\ell_\infty$. Defending against various attacks can be a challenging problem since multi-perturbation adversarial training and its variants only achieve suboptimal robustness trade-offs, due to the theoretical limit to multi-perturbation robustness for a single model. Besides, it is impractical to deploy large models in some storage-efficient scenarios. To settle down these drawbacks, in this paper we propose a novel multi-perturbation adversarial training framework, parameter-saving adversarial training (PSAT), to reinforce multi-perturbation robustness with an advantageous side effect of saving parameters, which leverages hypernetworks to train specialized models against a single perturbation and aggregate these specialized models to defend against multiple perturbations. Eventually, we extensively evaluate and compare our proposed method with state-of-the-art single/multi-perturbation robust methods against various latest attack methods on different datasets, showing the robustness superiority and parameter efficiency of our proposed method, e.g., for the CIFAR-10 dataset with ResNet-50 as the backbone, PSAT saves approximately 80\% of parameters with achieving the state-of-the-art robustness trade-off accuracy.
\end{abstract}

%%%%%%%%% BODY TEXT

\section{Introduction}\label{sec:intro}

In recent decades, deep neural networks (DNNs) have achieved mightily impressive success on various fields: computer vision \cite{Yu2022,Wang2022,Xu2023}, natural language processing \cite{Dao2022}, speech recognition \cite{Tueske2021}, graphs \cite{Wang2022a}. Nevertheless, we need to consider more about the robustness and stability than the precision when deploying these DNNs to real-world applications, or we will pay a heavy price for the unsafe application deployment. Particularly, it is evident that almost all DNNs are susceptible to \emph{adversarial examples} \cite{Szegedy2014}, i.e., samples that can be adversarially perturbed to mislead DNNs, but are very close to the original examples to be imperceptible to the human visual system. Typically, attackers generate adversarial examples by relying on gradient information to maximize the loss within a small perturbation neighbourhood, which is usually referred to as the adversary’s perturbation model.

How to defend against such adversarial examples has attracted remarkable attention of deep learning researchers. Lots of heuristic defences have been proposed, e.g., adversarial training \cite{Madry2018}, activation pruning \cite{Dhillon2018},  loss modification \cite{Pang2020}. Besides, some scholars improved the robustness of DNNs with provable methods \cite{Wong2018,Cohen2019,Zhang2023}. Nonetheless, despite massive literature to circumvent the above phenomenon, most previous works achieved the robustness with the hypothesis of a single adversarial perturbation. Therefore, it is quite challenging to achieve an optimal trade-off which minimizes the worst-case loss against multiple perturbations in practice. In other words, these methods converge to suboptimal local minima, accounting for a model that is highly robust to certain perturbations while failing to defend against others, and the robust performance often varies considerably across datasets, which causes unpredictable and inferior robust performance.

Moreover, some recent works \cite{Schott2018,Tramer2019,Maini2020,Madaan2021} have proposed some strategies to train a robust model against multiple perturbations. Notwithstanding, the existing works tried to train an all-round robust model, which is inferior to the single standard adversarially trained model for a certain perturbation. Due to the theoretical limits to multi-perturbation robustness for a single model (cf., Theorem 1 in \cite{Tramer2019}), a single model can only achieve a sub-optimal multi-perturbation robustness trade-off. It motivates us to leverage aggregated models to achieve high multi-perturbation robustness trade-off. However, directly aggregating single-perturbation robust models improves the robustness trade-off, while it also incurs a linear increase in the number of parameters, which may be unaffordable in some storage-efficient settings. 

To improve the multi-perturbation robustness trade-off as well as reduce the number of model parameters, we propose a novel adversarial training framework, parameter-saving adversarial training (PSAT), against multiple perturbations, which leverages the idea of hypernetwork to train specialized models against a single perturbation and aggregate these specialized models to defend against multiple perturbations. Specifically, we first design a parameter-saving hypernetwork backbone for some classifier, and adversarially train specialized hypernetworks for each perturbation type. Then, we aggregate these hypernetworks to defend against multiple perturbations. Note that even though it seems the aggregated model contains much more parameters than a standard classifier, the number of parameters of it is in fact much fewer than that of a standard classifier, since the hypernetwork backbone saves very many parameters compared with a standard classifier. Last but no least, we utilize lowest entropy strategy to select the optimal hypernetwork in the inference stage, reinforcing the multi-perturbation robustness trade-off.

Furthermore, we extensively demonstrate the multi-perturbation robustness superiority and parameter  efficiency of our proposed method by evaluating it on latest attack methods and comparing it with existing state-of-the-art single-perturbation and multi-perturbation robust methods on several benchmark datasets (CIFAR-10, SVHN, and TinyImageNet). The experimental results show that our proposed method achieves substantially superior performance over all the counterparts trained with multiple adversarial perturbations. Besides, our proposed method can generalize to diverse perturbations well and significantly reduce the parameter number. We summarize the contributions of this paper as follows:
\begin{itemize}
  \item We design a parameter-saving hypernetwork backbone for some classifier against a single adversarial perturbation.
  \item We aggregate the proposed specialized hypernetworks to defend against multiple perturbations and utilize lowest entropy strategy to select the final output of the aggregated model for each input.
  \item We evaluate our proposed method on various datasets against diverse adversarial attacks, on which it obtains state-of-the-art performance compared with latest counterpart methods.
\end{itemize} 

% The following sections are organized as below. Sections \ref{sec:related_work} and \ref{sec:preliminaries} introduce recent related works as well as the preliminary algorithms of our setting. Then, we illustrate our proposed parameter-saving adversarial training robust method in Section \ref{sec:method}. Besides, Section \ref{sec:exp} showcases extensive experimental results. Eventually, Sections \ref{sec:conclusion} concludes this paper.
\section{Related Work}\label{sec:related_work}

Since proposed by Goodfellow {\it et al.} \cite{Goodfellow2015}, there has been a plethora of papers studying the adversrial robustness, ranging from heuristic defences to certified defenses. In this section, we highlight a few related works that are most relevant to our setting.

% Since proposed by Goodfellow {\it et al.} \cite{Goodfellow2015}, many of the works have focused on how to defend against such single-perturbation adversarial threat models. Since then, there has been a plethora of papers studying the defence methods, ranging from heuristic defences to certified defenses. As there are far too many to discuss here, we highlight a few which are the most relevant to our setting in this section.

\noindent\textbf{Single-perturbation robustness.}\quad
Numerous methods were proposed to defend against a single perturbation, such as \cite{Dhillon2018,Madry2018,Carmon2019,Wu2020}, among which \emph{adversarial training} \cite{Madry2018} and its variants \cite{Zhang2019,Wang2019,Stutz2020} are one of most popular and most effective methods, which aim to train a surrogate model with adversarial examples generated by projected gradient descent (PGD) \cite{Madry2018} with some norm, and then uses this surrogate model to defend against adversarial attacks. They remain quite popular since it continues to perform well in various empirical benchmarks, though it comes with no formal guarantees. More recently, some variants have been proposed to further improve the robustness performance, like input transform \cite{Guo2017,Xie2018,Li2021}, revising loss functions \cite{Wang2020,Sriramanan2020,Pang2020}, adversarial data augmentation \cite{Wang2021,Rebuffi2021}, provable defenses \cite{Cohen2019,Lecuyer2019,Zhang2023}. However, most of these works focused primarily on defending against and verifying only a single adversarial perturbation. When extending them to multiple perturbations, the robustness is unpredictable or inferior.

\noindent\textbf{Multi-perturbation robustness.}\quad
Some works tried to settle down the multi-perturbation robustness problem. For instance, Schott {\it et al.} \cite{Schott2018} developed a complicated variational autoencoders (VAEs) based complicated framework, analysis by synthesis (ABS), for the MNIST dataset to withstand $\ell_\infty$, $\ell_2$ and $\ell_0$ attacks. Nevertheless, it was not scalable and limited to the MNIST dataset. Besides, Kang and his co-authors \cite{Kang2019} studied the robustness transferability against different perturbations; Croce and Hein \cite{Croce2020} presented a provable defence method against all $\ell_p$ norms for $p\ge 1$ via a regularization term. Particularly, Tram\`{e}r and Boneh \cite{Tramer2019} analyzed the theoretical and empirical robustness trade-offs in various settings when defending against aggregations of multiple adversaries and proposed two effective strategies, ``max'' and ``average'', to improve multi-perturbation robustness, while Maini {\it et al.} \cite{Maini2020} further considered multiple perturbations in every PGD step to train multi-perturbation robust models. Additionally, Madaan {\it et al.} \cite{Madaan2021} leveraged the generative method and the meta learning idea to speed up multi-perturbation adversarial training with the cost of increasing parameter number. These defence methods can be seen to establish an all-round model to defend against multiple perturbations by sacrificing the accuracy against a specific perturbation compared with single-perturbation robust models. Instead, in this paper we try to enhance multi-perturbation robustness with hypernetwork, bringing a beneficial  effect of reducing model parameters.

\noindent\textbf{Hypernetworks.}\quad
Proposed by Ha {\it et al.} \cite{Ha2017}, hypernetworks are aimed to use a small network to generate the weights for a larger network (denoted as a main network). The performance the the generated network is usually the same as that of the main network with direct optimization: hypernetworks also learn to map some raw data to their desired targets, while hypernetworks take a set of embeddings that contain information about the structure of the weights and generates the weight for that layer. These switch-like pocket networks show superiority in some computer vision tasks:  Oswald {\it et al.} \cite{Oswald2020} utilized hypernetworks to alleviate the catastrophic forgetting phenomenon in the continual learning task; Alaluf {\it et al.} \cite{Alaluf2022} and Dinh {\it et al.} \cite{Dinh2022} simultaneously used hypernetworks to improve image editing; Peng {\it et al.} \cite{Peng2022} applied hypernetworks to the task of 3D medical image segmentation. In this paper, we extend the hypernetwork framework to multi-perturbation adversarial robustness.
\section{Preliminaries}\label{sec:preliminaries}

%Before introducing our method, we first briefly introduce necessary preliminaries related to our setting: adversarial training and multiple attack defence.

\subsection{Adversarial Training}
Proposed by Madry {\it et al.} \cite{Madry2018}, adversarial training (AT) is one of the most effective approach to defend against adversarial attack, which minimizes the worst-case loss within certain perturbation range. Concretely, given training set $\{x_i,y_i\}_{i=1,\cdots,n}$, some classifier $f_\theta$ parameterized by $\theta$, some loss function $\mathcal{L}$ (e.g., cross-entropy loss), the formulation of adversarial training is 
\begin{equation}\label{eqn:at}
\small
\begin{aligned}
    & \min_\theta\frac{1}{n}\sum_{i=1}^n\mathcal{L}(f_\theta(x_i+\delta_p),y_i), \\
    & \text{s.t.~~}\delta_p=\mathop{\mathrm{argmax}}\limits_{\delta_p\in\mathcal{B}(p,\epsilon)}\mathcal{L}(f_\theta(x_i+\delta_p),y_i),\\
\end{aligned}
\end{equation}
where $\mathcal{B}(p,\epsilon)=\{\delta_p:\|\delta_p\|_p\le\epsilon\}$ is the $\ell_p$ ball with radius $\epsilon$ centered at $x_i$; $\delta$ denotes the worst-case adversarial perturbations. We usually utilize PGD to solve the inner maximization. For $\tau$-step PGD, we first define the projection operation onto $\mathcal{B}(p,\epsilon)$ to be 
\begin{equation}
\small
\mathop{\mathrm{proj}}\limits_{\mathcal{B}(p,\epsilon)}(\omega)=\mathop{\mathrm{argmax}}\limits_{s\in\mathcal{B}(p,\epsilon)}\|\omega-s\|^2_2,
\end{equation}
which searches the point closest to the input $\omega$ that lies within the ball $\mathcal{B}(p,\epsilon)$ in Euclidean space. Then, for some step size $\alpha$, the algorithm includes the below iteration:
\begin{equation}
\small
\delta_p^{t+1}=\mathop{\mathrm{proj}}\limits_{\mathcal{B}(p,\epsilon)}\left(\delta_p^{t}+\mathop{\mathrm{argmax}}\limits_{\|v\|_p\le\alpha}v^{T}\nabla\mathcal{L}(f_\theta(x_i+\delta_p^{t}),y_i)\right),
\end{equation}
where $t=0,1,\cdots,\tau-1$; $\delta_p^0=0~\text{or random value}$. 
%This is sometimes called as \emph{projected steepest descent}, which is used to generated adversarial examples since the standard gradient steps are typically too small.

%------------------------------------------------------------------------------------------------
\subsection{Multi-Perturbation Adversarial Training}
Normally, Eqn. (\ref{eqn:at}) is against a single perturbation, which is usually called as standard adversarial training. In this paper, we mainly consider how adversarial training defends against multiple perturbations. Formally, let $\mathcal{A}$ denotes a set of adversarial perturbations, such taht $p\in\mathcal{A}$ corresponds to the $\ell_p$ perturbation ball $\mathcal{B}(p,\epsilon)$, and let $\mathcal{B}(\mathcal{A},\epsilon)=\bigcup_{p\in\mathcal{A}}\mathcal{B}(p,\epsilon)$ represent the union of all perturbations in the set $\mathcal{A}$. Although the radii $\epsilon$ are different in different perturbations, we still use the same symbol $\epsilon$ for brevity. Thus, the version that defends against multiple perturbations of Eqn. (\ref{eqn:at}) is formulated as
\begin{equation}\label{eqn:mul_at}
\small
\min_\theta\frac{1}{n}\sum_{i=1}^n\mathcal{L}(f_\theta(x_i+\delta_\mathcal{A}),y_i).
\end{equation}
Here, previous works proposed several different methods to generate $\delta_\mathcal{A}$ or optimize (\ref{eqn:mul_at}).

\noindent\textbf{Worst-case manner.}\quad
One simple way to generate $\delta_\mathcal{A}$ is to select the most aggressive perturbation among the perturbation set $\mathcal{A}$. Concretely, for each perturbation $p\in\mathcal{A}$, we approximate the worst-case perturbation by solving
\begin{equation}\label{eqn:max_at}
\small
\delta_\mathcal{A}=\mathop{\mathrm{argmax}}\limits_{\delta_\mathcal{A}\in\mathcal{B}(\mathcal{A},\epsilon)}\mathcal{L}(f_\theta(x_i+\delta_\mathcal{A}),y_i).
\end{equation}
Here, if $|\mathcal{A}|=1$, it degrades into standard adversarial training (\ref{eqn:at}). This manner is called ``max'' strategy.

\noindent\textbf{Average manner.}\quad
Another direct approach to adversarially train multiple robust classifiers is to use the average loss of all perturbations in the set $\mathcal{A}$ for adversarial training. Specifically, we optimize (\ref{eqn:mul_at}) by 
\begin{equation}\label{eqn:avg_at}
\small
\min_\theta\frac{1}{n}\sum_i^n\left[\frac{1}{|\mathcal{A}|}\sum_{p}^{|\mathcal{A}|}\max_{\delta_{p}\in\mathcal{B}(p,\epsilon)}\mathcal{L}(f_\theta(x_i+\delta_{p}),y_i)\right].
\end{equation}
Similarly, the above optimization will degrade into standard adversarial training when $|\mathcal{A}|=1$. This manner is called ``avg'' strategy.

%------------------------------------------------------------------------------------------------
\noindent\textbf{Multi steepest descent manner.}\quad
Maini {\it et al.} \cite{Maini2020} refined the worst-case manner (\ref{eqn:max_at}) by approximating the worst-case perturbation in every PGD step:
\begin{equation}
\small
\begin{aligned}
&\delta^{t+1}_\mathcal{A}=\mathop{\mathrm{proj}}\limits_{\mathcal{B}(p,\epsilon)}\left(\delta^{t}_\mathcal{A}+\Delta^t_\mathcal{A}\right),\\
&\text{s.t.~~}\Delta^t_\mathcal{A}=\mathop{\mathrm{argmax}}\limits_{\delta^t_\mathcal{A}\in\mathcal{B}(\mathcal{A},\epsilon)}\mathcal{L}(f_\theta(x_i+\delta^t_\mathcal{A}),y_i),\\
\end{aligned}
\end{equation}
where $t=0,\cdots,\tau-1$.

\noindent\textbf{Noise learning manner.}\quad
Madaan {\it et al.} \cite{Madaan2021} took advantage of the idea of meta learning to make all perturbations consistent and utilized stochastic adversarial training to save training time. Nevertheless, such manner increases the number of parameters, since the authors added an auxiliary network (the meta-noise generator) to the training stage.
\section{Parameter-Saving Adversarial Training}\label{sec:method}

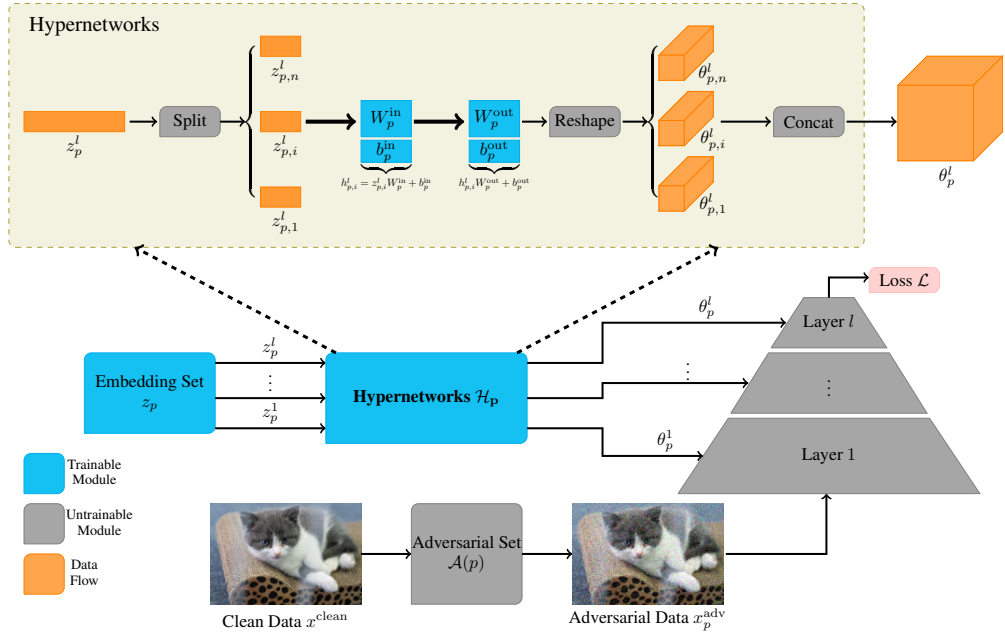
\begin{figure*}[t]
\centering
\resizebox{0.75\linewidth}{!}{
    \begin{tikzpicture}
        %%变量
\newcommand{\xBias}{0}
\newcommand{\yBias}{0}
\newcommand{\xxBias}{0}
\newcommand{\yyBias}{0}
\newcommand{\scaling}{0}

%%函数
%多边形
\newcommand{\arrowWidth}{0.04cm}
\newcommand{\mypolygon}[8]{
%{#1:x}{#2:y}{#3:h}{#4:l}{#5:w}{#6:color}{#7:opacity}{#8:rounded corners}
    \draw[#6!100, fill=#6!#7, #8] (#1-0.5*#4,#2) -- (#1+0.5*#4,#2) -- (#1+0.5*#5,#2+#3) -- (#1-0.5*#5,#2+#3) -- (#1-0.5*#4,#2);
}

%长方体
\newcommand{\mycuboid}[9]{
%{#1:Ox}{#2:Oy}{#3:Oz}{#4:l}{#5:w}{#6:h}{#7:color}{#8:line opacity}{#9:fill opacity}
\coordinate (O) at (#1,#2,#3);
\coordinate (A) at (#1,#2,#3+#6);
\coordinate (B) at (#1+#4,#2,#3+#6);
\coordinate (C) at (#1+#4,#2,#3);
\coordinate (D) at (#1,#2+#5,#3);
\coordinate (E) at (#1,#2+#5,#3+#6);
\coordinate (F) at (#1+#4,#2+#5,#3+#6);
\coordinate (G) at (#1+#4,#2+#5,#3);
\draw[#7!#8,fill=#7!#9] (O) -- (A) -- (B) -- (C) -- cycle;% Front Face
\draw[#7!#8,fill=#7!#9] (O) -- (A) -- (E) -- (D) -- cycle;% Left Face
\draw[#7!#8,fill=#7!#9] (O) -- (C) -- (G) -- (D) -- cycle;% Bottom Face
\draw[#7!#8,fill=#7!#9] (D) -- (E) -- (F) -- (G) -- cycle;% Back Face
\draw[#7!#8,fill=#7!#9] (B) -- (C) -- (G) -- (F) -- cycle;% Right Face
\draw[#7!#8,fill=#7!#9] (A) -- (B) -- (F) -- (E) -- cycle;% Top Face
}

\renewcommand{\xBias}{3.2}
\renewcommand{\yBias}{1.3}
\node at (0+\xBias,0+\yBias) {\includegraphics[width=3cm,height=2cm]{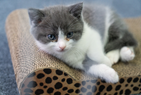}};
\node at (0+\xBias,-1.3+\yBias) {Clean Data $x^\mathrm{clean}$};
\draw[->,line width=\arrowWidth] (1.5+\xBias,0+\yBias) -- +(1,0);
\mypolygon{3+2.2/2-0.5+\xBias}{-1+\yBias}{2}{2.2}{2.2}{gray}{70}{rounded corners};
\node at (3+2.2/2-0.5+\xBias,0+\yBias) {\begin{tabular}{c}Adversarial Set\\ $\mathcal{A}(p)$\\\end{tabular}};
\draw[->,line width=\arrowWidth] (3+2.2-0.5+\xBias,0+\yBias) -- +(1,0);
\node at (3+2.2+1.5+0.5+\xBias,\yBias) {\includegraphics[width=3cm,height=2cm]{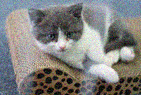}};
\node at (3+2.2+1.5+0.5+\xBias,-1.3+\yBias) {Adversarial Data $x_p^\mathrm{adv}$};
\draw[->,line width=\arrowWidth] (3+2.2+3+0.55+\xBias,\yBias) -- ++(2,0) -- +(0,1.2);

\renewcommand{\xBias}{0}
\renewcommand{\yBias}{-1.5}

\mypolygon{14+\xBias}{4+\yBias}{1.5}{6}{4}{gray}{70}{};
\node at (14+\xBias,4+1.5*0.5+\yBias) {Layer $1$};
\mypolygon{14+\xBias}{5.6+\yBias}{1.2}{3.9}{2.4}{gray}{70}{};
\node at (14+\xBias,5.6+1.2/2+\yBias) {$\vdots$};
\mypolygon{14+\xBias}{6.9+\yBias}{1}{2.3}{1}{gray}{70}{};
\node at (14+\xBias,6.9+1/2+\yBias) {Layer $l$};
\draw[->,line width=\arrowWidth] (14+\xBias,7.9+\yBias) -- ++(0,0.4) -- +(0.8,0);
\mypolygon{15.5+\xBias}{8+\yBias}{0.5}{1.4}{1.4}{pink}{70}{rounded corners};
\node at (15.5+\xBias,8+0.5/2+\yBias) {Loss $\mathcal{L}$};

\renewcommand{\xxBias}{0}
\renewcommand{\yyBias}{-1.5}
\mypolygon{0.5+\xxBias}{5.2+\yyBias}{1.6}{2.6}{2.6}{cyan}{70}{rounded corners};
\node at (0.5+\xxBias,5.2+1.6/2+\yyBias) {\begin{tabular}{c}Embedding Set\\$z_p$\\\end{tabular}};
\draw[->,line width=\arrowWidth] (1.8+\xxBias,6.6+\yyBias) -- (4+\xxBias,6.6+\yyBias);
\node at (1.8/2+4/2+\xxBias,6.9+\yyBias) {$z^l_p$};
\draw[->,line width=\arrowWidth] (1.8+\xxBias,5.9+\yyBias) -- (4+\xxBias,5.9+\yyBias);
\node at (1.8/2+4/2+\xxBias,6.3+\yyBias) {$\vdots$};
\draw[->,line width=\arrowWidth] (1.8+\xxBias,5.3+\yyBias) -- (4+\xxBias,5.3+\yyBias);
\node at (1.8/2+4/2+\xxBias,5.6+\yyBias) {$z^1_p$};
\mypolygon{6+\xxBias}{5+\yyBias}{1.8}{4}{4}{cyan}{70}{rounded corners};
\node at (6+\xxBias,5+1.8/2+\yyBias) {\textbf{Hypernetworks} $\mathbf{\mathcal{H}_p}$};

\draw[->,line width=\arrowWidth] (8+\xxBias,6.6+\yyBias) -- (8+1.5+\xxBias,6.6+\yyBias) -- (8+1.5+\xBias,6.9+1/2+\yBias) -- (13.15+\xBias,6.9+1/2+\yBias);
\node at (10/2+13.15/2+\xxBias/2+\xBias/2,6.9+1/2+0.3+\yBias/2++\yyBias/2) {$\theta^l_p$};
\draw[->,line width=\arrowWidth] (8+\xxBias,5.9+\yyBias) -- (9.5+\xxBias,5.9+\yyBias) -- (9.5+\xBias,5.6+1.2/2+\yBias) -- (12.4+\xBias,5.6+1.2/2+\yBias);
\node at (10/2+12.4/2+\xxBias/2+\xBias/2,5.6+1.2/2+0.33+\yBias/2++\yyBias/2) {$\vdots$};
\draw[->,line width=\arrowWidth] (8+\xxBias,5.3+\yyBias) -- (9.5+\xxBias,5.3+\yyBias) -- (9.5+\xBias,4+1.5*0.5+\yBias) -- (11.5+\xBias,4+1.5*0.5+\yBias);
\node at (10/2+11.5/2+\xxBias/2+\xBias/2,4+1.5*0.5+0.3+\yBias/2++\yyBias/2) {$\theta^1_p$};

%-------------------------------------------------------------------------------------

\renewcommand{\xBias}{-1}
\renewcommand{\yBias}{9.7}
\draw[rounded corners,olive!80,dashed,fill=olive!10] (\xBias-1.3,\yBias-2.3) -- +(17,0) -- +(17,4.85) -- +(0,4.85) --cycle;
\node at (\xBias+0.4,\yBias+-2.3+4.4)[scale=1.2] {Hypernetworks};
\draw[->,line width=0.06cm,dashed] (\xBias+5.2,\yBias-4.4) -- +(-4,2.1);
\draw[->,line width=0.06cm,dashed] (\xBias+8.8,\yBias-4.4) -- +(4,2.1);
\mypolygon{\xBias}{\yBias}{0.4}{2}{2}{orange}{80}{};
\node at (\xBias,-0.3+\yBias) {$z^l_p$};
\draw[->,line width=\arrowWidth] (1.1+\xBias,0.2+\yBias) -- +(0.55,0);
\mypolygon{\xBias+2.3}{\yBias-0.1}{0.6}{1.2}{1.2}{gray}{70}{rounded corners};
\node at (\xBias+2.3,\yBias+0.2) {Split};
\draw[->,line width=\arrowWidth] (\xBias+2.9,0.2+\yBias) -- +(0.5,0);
\node[rotate = 270] at (\xBias+3.5, \yBias+0.2) {$\underbrace{\hspace{3cm}}$};
\mypolygon{\xBias+4.1}{\yBias}{0.4}{0.8}{0.8}{orange}{80}{};
\node at (\xBias+4.2,\yBias-0.3) {$z^l_{p,i}$};
\mypolygon{\xBias+4.1}{\yBias-1.5}{0.4}{0.8}{0.8}{orange}{80}{};
\node at (\xBias+4.2,\yBias-1.5-0.3) {$z^l_{p,1}$};
\mypolygon{\xBias+4.1}{\yBias+1.5}{0.4}{0.8}{0.8}{orange}{80}{};
\node at (\xBias+4.2,\yBias+1.5-0.3) {$z^l_{p,n}$};
\mypolygon{\xBias+6.2}{\yBias-0.1}{0.72}{1}{1}{cyan}{70}{};
\node at (\xBias+6.2,\yBias-0.1+0.72/2) {$W^\mathrm{in}_p$};
\mypolygon{\xBias+6.2}{\yBias-0.51-0.1}{0.44}{1}{1}{cyan}{70}{};
\node at (\xBias+6.2,\yBias-0.51+0.44/2-0.1) {$b^\mathrm{in}_p$};
\draw[->,line width=0.08cm] (\xBias+4.6,\yBias+0.2) -- +(1,0);
\node[rotate = 0] at (\xBias+6.2, \yBias-.64-0.1) {$\underbrace{\hspace{1cm}}$};
\node at (\xBias+6.2, \yBias-.64-0.3-0.1)[scale=0.6] {$h^l_{p,i}=z^l_{p,i}W^\mathrm{in}_p+b^\mathrm{in}_p$};
\draw[->,line width=0.08cm] (\xBias+6.75,0.2+\yBias) -- +(1,0);
\mypolygon{\xBias+6.75+1.6}{\yBias-0.1}{0.72}{1}{1}{cyan}{70}{};
\node at (\xBias+6.75+1.6,\yBias-0.1+0.72/2) {$W^\mathrm{out}_p$};
\mypolygon{\xBias+6.75+1.6}{\yBias-0.51-0.1}{0.44}{1}{1}{cyan}{70}{};
\node at (\xBias+6.75+1.6,\yBias-0.51+0.44/2-0.1) {$b^\mathrm{out}_p$};
\node[rotate = 0] at (\xBias+\xBias+6.75+1.6+1, \yBias-.64-0.1) {$\underbrace{\hspace{1cm}}$};
\node at (\xBias+\xBias+6.75+1.6+1, \yBias-.64-0.3-0.1)[scale=0.6] {$h^l_{p,i}W^\mathrm{out}_p+b^\mathrm{out}_p$};
\draw[->,line width=\arrowWidth] (\xBias+6.75+2.15,0.2+\yBias) -- +(0.5,0);
\mypolygon{\xBias+6.75+2.15+0.5+0.7+0.05}{\yBias-0.1}{0.6}{1.4}{1.4}{gray}{70}{rounded corners};
\node at (\xBias+6.75+2.15+0.5+0.7+0.05,\yBias+0.2) {Reshape};
\draw[->,line width=\arrowWidth] (\xBias+6.75+2.15+0.5+1.4+0.1,0.2+\yBias) -- +(0.5,0);
\node[rotate = 270] at (\xBias+6.75+2.15+0.5+1.4+0.1+0.5+0.1,0.2+\yBias) {$\underbrace{\hspace{3cm}}$};
\mycuboid{\xBias+6.75+2.15+0.5+1.4+0.1+0.5+0.1+0.7}{\yBias+1.6}{0}{0.5}{0.5}{1.5}{orange}{100}{70};
\node at (\xBias+6.75+2.15+0.5+1.4+0.1+0.5+0.1+1.2,\yBias+1.6-0.45) {$\theta^l_{p,n}$};
\mycuboid{\xBias+6.75+2.15+0.5+1.4+0.1+0.5+0.1+0.7}{\yBias+0.3}{0}{0.5}{0.5}{1.5}{orange}{100}{70};
\node at (\xBias+6.75+2.15+0.5+1.4+0.1+0.5+0.1+1.2,\yBias+0.3-0.45) {$\theta^l_{p,i}$};
\mycuboid{\xBias+6.75+2.15+0.5+1.4+0.1+0.5+0.1+0.7}{\yBias-1}{0}{0.5}{0.5}{1.5}{orange}{100}{70};
\node at (\xBias+6.75+2.15+0.5+1.4+0.1+0.5+0.1+1.2,\yBias-1-0.45) {$\theta^l_{p,1}$};
\draw[->,line width=\arrowWidth] (\xBias+6.75+2.15+0.5+1.4+0.1+0.5+0.1+1.2+0.15,0.2+\yBias) -- +(1,0);
\mypolygon{\xBias+6.75+2.15+0.5+1.4+0.1+0.5+0.1+1.2+0.15+1+0.75}{\yBias-0.1}{0.6}{1.4}{1.4}{gray}{70}{rounded corners};
\node at (\xBias+6.75+2.15+0.5+1.4+0.1+0.5+0.1+1.2+0.15+1+0.75,\yBias+0.2) {Concat};
\draw[->,line width=\arrowWidth] (\xBias+6.75+2.15+0.5+1.4+0.1+0.5+0.1+1.2+0.15+1+1.4+0.1,0.2+\yBias) -- +(1,0);
\mycuboid{\xBias+6.75+2.15+0.5+1.4+0.1+0.5+0.1+1.2+0.15+1+1.4+1+0.7}{\yBias}{0}{1.5}{1.5}{1.5}{orange}{100}{70};
\node at (\xBias+6.75+2.15+0.5+1.4+0.1+0.5+0.1+1.2+0.15+1+1.4+1+1.1,\yBias-0.9) {$\theta^l_p$};

\renewcommand{\xBias}{-1.6}
\renewcommand{\yBias}{2.5}
\mypolygon{\xBias}{\yBias}{0.8}{0.8}{0.8}{cyan}{70}{rounded corners};
\node at (\xBias+1,\yBias+0.4) [scale=0.8]{\begin{tabular}{c}Trainable\\Module\end{tabular}};
\mypolygon{\xBias}{\yBias-1}{0.8}{0.8}{0.8}{gray}{70}{rounded corners};
\node at (\xBias+1.1,\yBias-0.6) [scale=0.8]{\begin{tabular}{c}Untrainable\\Module\end{tabular}};
\mypolygon{\xBias}{\yBias-2}{0.8}{0.8}{0.8}{orange}{70}{rounded corners};
\node at (\xBias+0.8,\yBias-1-0.6)[scale=0.8] {\begin{tabular}{c}Data\\Flow\end{tabular}};
   \end{tikzpicture}
}
\vspace{-0.2cm}
\caption{Overview of our proposed method. We leverage the embedding set and the hypernetworks to generate parameters for the main network. Note that the main network is only a forward propagation network without optimization, while we optimize the embedding set and the hypernetworks.}
\label{fig:train}
\vspace{-0.5cm}
\end{figure*}

%In this section, we present our parameter-saving adversarial training (PSAT) framework against multiple perturbations, which utilizes hypernetworks to reduce the number of network parameters with persevering robustness against multiple perturbations.

According to the theorems in \cite{Tramer2019}, there exist theoretical limits to multi-perturbation robustness for a single model. For example, a $\ell_\infty$ robust model cannot be a good robust model against $\ell_1$ perturbation (cf., Theorem 1 in \cite{Tramer2019}). From Table \ref{tab.main_results}, we can find that there exists a remarkable adversarial accuracy gap for a single perturbation between the single-perturbation robust models and the multi-perturbation robust models. It is reasonable that the multi-perturbation models are inferior to the specialized model for a certain adversarial perturbation. Thus, our motivation is to avoid letting one model combat too many adversarial attacks. Instead, we opt to have a robust model defend against a corresponding attack and aggregate these specialists to reinforce the multi-perturbation robustness trade-off. Here, we wonder: \emph{Can we become specialists that specialize in multiple perturbations?} A direct idea is to train many specialized models against all the perturbations. Nonetheless, this way will increase linearly the number of parameters, e.g., for $N$ perturbations, the parameter number of the final aggregated model is $N$ times more than that of a single model (cf., Table \ref{tab.direct_aggregate}). Sometimes, such big model is unbearable in some storage-efficient setting. In this paper, we solve the above two problems via hypernetworks.

%------------------------------------------------------------------------------------------------
\subsection{Overview}
For some perturbation (denoted as $p$-perturbation), instead of directly fine-tuning the parameters $\theta$ of a standard classifier $f_\theta$, we train a hypernetwork $\mathcal{H}_p$ to generate parameters of each layer of the target classifier $f_\theta$, which can be formulated as
\begin{equation}\label{eqn:hyper}
\small
\theta^l_p = \mathcal{H}_p(z^l_p),
\end{equation}
where $l$ is the $l$-layer of the classifier $f_\theta$; $z^l$ denotes the input embedding of the hypernetwork $\mathcal{H}$ for generating parameters of the $l$-layer; the subscript $p$ means the $p$-perturbation. Here, hypernetwork is sometimes referred to weight generator. Thus, the hypernetwork version of adversarial training against a single $p$-perturbation can be rewritten as
\begin{equation}
\small
\begin{aligned}
&\min_{\mathcal{H}_p,z_p}\frac{1}{n}\sum_{i=1}^n\mathcal{L}(f_{\mathcal{H}_p(z_p)}(x_i+\delta_p),y_i), \\
&\text{~~s.t.~}\delta_p=\mathop{\mathrm{argmax}}\limits_{\delta_p\in\mathcal{B}(p,\epsilon)}\mathcal{L}(f_{\mathcal{H}_p(z_p)}(x_i+\delta_p),y_i),\\
\end{aligned}
\end{equation}
where $z_p=\{z^1_p,\cdots,z^l_p,\cdots\}$ is the total embeddings of all layers.

For multiple perturbations, we train multiple hypernetworks as well as corresponding embeddings, and then integrate them as the aggregated model $\mathcal{F}_\mathcal{H}$ that generalizes to all the perturbations, which can be formulated as
\begin{equation}\label{eqn:psat}
\small
\begin{aligned}
&\mathcal{F}_\mathcal{H}=\bigcup_{p\in\mathcal{A}}\{\mathcal{H}_p,z_p\}, \\
&\text{s.t.~}\mathcal{H}_p,z_p=\mathop{\mathrm{argmin}}\limits_{\mathcal{H}_p,z_p}\mathcal{L}(f_{\mathcal{H}_p(z_p)}(x_i+\delta_p),y_i),\\
&\text{~~~~~~}\delta_p=\mathop{\mathrm{argmax}}\limits_{\delta_p\in\mathcal{B}(p,\epsilon)}\mathcal{L}(f_{\mathcal{H}_p(z_p)}(x_i+\delta_p),y_i).\\
\end{aligned}
\end{equation}
Even though the aggregated model (\ref{eqn:psat}) looks like it has lots of parameters, it saves many parameters when compared with previous multiple robust models, since every hypernetwork in $\mathcal{F}_\mathcal{H}$ contains much fewer parameters than those of the original model (quantitative experimental results will be displayed in the Experiments section). Therefore, the proposed method is referred to \underline{P}arameter-\underline{S}aving \underline{A}dversarial \underline{T}raining (PSAT) against multiple perturbations. We summarize our proposed method in Figure \ref{fig:train} and the algorithm details are shown in the supplementary material.

\begin{table*}[t]
\centering
\caption{Clean and adversarial accuracy against multiple perturbations of our proposed PSAT model and counterparts for CIFAR-10, SVHN and TinyImageNet datasets. For CIFAR-10 and SVHN datasets, we used ResNet-50 as the basic network architecture; for TinyImageNet dataset, we used ResNet-18 as the basic network architecture. The best accuracy are highlighted in \textbf{bold} for every evaluated attack (every column).}
\vspace{-0.2cm}
\resizebox{\linewidth}{!}{
\begin{tabular}{clcccccccrc}
\toprule[1pt]
Dataset & Method & Acc$_\text{clean}$     & Acc$_{\ell_\infty}$    & Acc$_{\ell_2}$      & Acc$_{\ell_1}$      & Acc$^\text{max}_\text{adv}$     & Acc$^\text{avg}_\text{adv}$      & Acc$_\text{trade-off}$ & \# of Paras &  Paras Saving\\
\cline{1-11}
\multirow{9}*{CIFAR-10} & NAT      & \textbf{92.82\%} & 0.00\%  & 0.32\%  & 0.11\%  & 0.00\%  & 0.14\%     & 15.57\%  & \multirow{7}*{23.547M} &\multirow{7}*{-}\\
\cline{2-9}
&AT$_{\ell_\infty}$     & 84.86\% & \textbf{42.80\%} & 53.97\% & 25.29\% & 24.23\% & 40.66\%    & 45.30\%  & &\\
&AT$_{\ell_2}$       & 87.18\% & 27.84\% & \textbf{63.96\%} & 52.76\% & 26.85\% & 48.18\%    & 51.13\%  & &\\
&AT$_{\ell_1}$       & 90.32\% & 0.59\%  & 1.46\%  & \textbf{78.11\%} & 0.00\%  & 26.72\%    & 32.87\%  & &\\
\cline{2-9}
&AT$_\text{max}$       & 82.54\% & 39.74\% & 62.46\% & 56.10\% & 39.26\% & 52.77\%    & 55.48\%  & &\\
&AT$_\text{avg}$       & 84.20\% & 34.12\% & 62.75\% & 60.42\% & 34.09\% & 52.45\%    & 54.67\%  & &\\
&AT$_\text{msd}$       & 79.08\% & 42.01\% & 60.89\% & 53.48\% & \textbf{42.22\%} & 52.11\%    & 54.97\%  &  &\\
\cline{2-11}
&AT$_\text{mng}$       & 77.92\% & 41.40\% & 62.64\% & 62.30\% & 41.31\% & 55.46\%    & 56.84\%  & 23.552M  & $\textcolor{red}{\uparrow}0.02\%$\\
\cline{2-11}
&PSAT (\textbf{Ours})     & 82.28\% & 40.33\% & 60.28\% & 68.13\% & 40.78\% & \textbf{56.32\%}    & \textbf{58.02\%}  & 4.874M  &$\textcolor{green}{\downarrow}79.30\%$ \\

\midrule[1pt]
\multirow{9}*{SVHN} & NAT                & \textbf{95.95\%} & 0.00\%           & 3.81\%           & 4.26\%           & 0.00\%           & 2.69\%  & 17.79\%     & \multirow{7}*{23.547M} & \multirow{7}*{-}\\
\cline{2-9}
& AT$_{\ell_\infty}$ & 92.60\%          & \textbf{44.62\%} & 32.93\%          & 12.26\%          & 10.18\%          & 29.92\% & 37.09\%     &    & \\
& AT$_{\ell_2}$      & 92.86\%          & 23.44\%          & \textbf{63.38\%} & 45.86\%          & 21.48\%          & 43.17\% & 48.37\%     &    & \\
& AT$_{\ell_1}$      & 92.21\%          & 0.12\%           & 0.00\%           & \textbf{75.03\%} & 0.00\%           & 25.08\% & 32.07\%     &    & \\
\cline{2-9}
& AT$_\text{max}$    & 90.26\%          & 28.50\%          & 55.27\%          & 50.43\%          & 28.15\%          & 44.71\% & 49.55\%     &    & \\
& AT$_\text{avg}$    & 91.83\%          & 22.40\%          & 55.62\%          & 60.75\%          & 19.71\%          & 46.24\% & 49.43\%     &    & \\
& AT$_\text{msd}$    & 83.17\%          & 33.45\%          & 53.43\%          & 44.15\%          & \textbf{33.60\%} & 43.68\% & 48.58\%     &    & \\
\cline{2-11}
& AT$_\text{mng}$ & 86.63\% & 34.04\% & 60.19\% & 66.40\% & 32.87\% & \textbf{54.62\%} & 55.79\% &  23.552M  & $\textcolor{red}{\uparrow}0.02\%$\\
\cline{2-11}
& PSAT (\textbf{Ours}) & 91.20\%          & 40.12\%          & 57.66\%          & 63.73\%          & 29.30\%          & 53.85\% & \textbf{55.98\%} & 4.874M  & $\textcolor{green}{\downarrow}79.30\%$\\
\midrule[1pt]

\multirow{9}*{TinyImageNet}&NAT    & \textbf{60.59\%} & 0.00\%           & 11.93\%          & 1.50\%           & 0.00\%           & 7.23\%           & 13.54\%      &  \multirow{7}*{11.276M}  & \multirow{7}*{-}\\
\cline{2-9}
&AT$_{\ell_\infty}$   & 53.95\%          & \textbf{28.62\%} & 40.56\%          & 33.53\%          & 27.53\%          & 34.21\%          & 36.40\%       &   &\\
&AT$_{\ell_2}$     & 58.95\%          & 7.80\%           & \textbf{42.31\%} & 47.44\%          & 6.81\%           & 32.52\%          & 32.64\%     &     &\\
&AT$_{\ell_1}$     & 56.66\%          & 9.19\%           & 40.38\%          & \textbf{48.65\%} & 9.23\%           & 32.72\%          & 32.81\%     &     &\\
\cline{2-9}
&AT$_\text{max}$    & 52.05\%          & 27.68\%          & 41.43\%          & 39.76\%          & 27.61\%          & 36.30\%          & 37.47\%      &    &\\
&AT$_\text{avg}$    & 55.64\%          & 20.06\%          & 41.41\%          & 46.92\%          & 22.04\%          & 36.14\%          & 37.04\%      &    &\\
&AT$_\text{msd}$     & 51.31\%          & 28.09\%          & 34.47\%          & 42.86\%          & \textbf{28.03\%} & 35.11\%          & 36.65\%      &    &\\
\cline{2-11}
&AT$_\text{mng}$     & 50.39\%          & 28.59\%          & 41.38\%          & 43.42\%          & 27.59\%          & \textbf{37.76\%} & 38.19\%       & 11.280M & $\textcolor{red}{\uparrow}0.04\%$  \\
\cline{2-11}
&PSAT (\textbf{Ours})   & 53.08\%          & 27.62\%          & 39.84\%          & 44.15\%          & 27.64\%          & 37.19\%          & \textbf{38.25\%}  & 1.318M &$\textcolor{green}{\downarrow}88.31\%$ \\
\bottomrule[1pt]
\end{tabular}
\label{tab.main_results}
}
\vspace{-0.4cm}
\end{table*}

\subsection{Hypernetwork Design}
The target classifier contains more than ten million parameters. On the one hand, controlling over too many parameters would account for an infeasible model that requires huge training resources. On the other hand, every layer contains different types of parameters. Thus, we need to design a unified approach to generate parameters of every layer, or the parameter number of hypernetwork will increases significantly.
Considering these two points, designing an expressive network is very challenging, which needs an elaborate balance between expressive performance and the selection as well as unification of generated parameters.

%------------------------------------------------------------------------------------------------
\noindent\textbf{Parameter selection.}\quad
%\paragraph{\textbf{Parameter selection.}}
Generally, for a standard CNN-based classifier, there are five primary types of layers: convolution layer, activation layer, batch normalization (BN) layer, pooling layer and fully connected layer. The activation layer and pooling layer do not contain any parameter. The widely-used BN layer includes two parameters, so it is unnecessary to utilize a complicated network to generate only two parameters. Besides, the fully connected layer usually contains thousands/hundreds times as many parameters as the number of total classes, which only accounts for a very small percentage in the total parameters. Thus, we will not generate the parameters of the fully connected layer. The last type of layer, convolution layer, contains almost all of the parameters in a model, which is the focus of our discussion. Usually, the parameter structure of some convolution layer $l$ is $C_l^\text{out}\times C_l^\text{in}\times k\times k$, where $k\times k$ is the size of convolution kernel; $C_l^\text{out}$ represents the total number of filters, each with $C_l^\text{in}$ channels.

%------------------------------------------------------------------------------------------------
\noindent\textbf{Generated parameter unification.}\quad
%\paragraph{\textbf{Generated parameter unification.}}
For popular networks (like ResNet), researchers prefer $3\times3$ convolution kernel and choose $64K (K=1,2,\cdots)$ channels/filters. Thus, we design the output unit of our hypernetwork is $64\times 64\times 3\times 3$ for unifying the structure of generated parameters. For parameters with larger structure, we concatenate the output unit in the filter $C_l^\text{out}$ and channel $C_l^\text{in}$ dimensions. For some simplified parameter structure, we downsample the kernel dimensions, e.g., use the average/maximum/sum operator to downsample $3\times 3$ kernel into $1\times 1$ kernel. Note that we do not generate the first convolution layer $64\times 3\times 3\times 3$, since the parameter number is very few compared with the total parameter number and it is not easy to generate them by the output unit.

%------------------------------------------------------------------------------------------------
\noindent\textbf{Hypernetwork structure.}\quad
%\paragraph{\textbf{Hypernetwork structure.}}
Inspired by \cite{Ha2017}, we design a two-layer linear network as our hypernetwork. The first layer takes the embedding $z^l$ as input and linearly projects it into the hidden layer. The second layer is a linear operation which takes the output of the hidden layer as input and linearly projects it into the ouput unit. Thus, the process that generates $l$-layer's parameters with hypernetwork can be written as
\begin{equation}
\small
\begin{aligned}
&h^l_{p,i}=W^{\text{in}}_pz^l_{p,i}+B^{\text{in}}_p,\\
&\theta^l_{p,i}=\langle W^{\text{out}}_p,h^l_{p,i}\rangle +B^{\text{out}}_p,\\
&\theta^l_p=\mathrm{concat}(\theta^l_{p,i})_{i=1,\cdots,D}\\
\end{aligned}
\end{equation}
where $D$ is the number of the output unit in the layer $l$; $z^l_p=\{z^l_{p,i}\}_{i=1,\cdots,N}$. Here the trainable parameters of hypernetwork $\mathcal{H}_p$ are $W^{\text{in}}_p$, 
$B^{\text{in}}_p$, $W^{\text{out}}$ and $B^{\text{out}}_p$. Additionally, the layer embeddings $z_p=\{z_p^l\}_{l=1,\cdots,L}$ ($L$ is the number of total generated convolution layers) are also learnable. Note that parameters of different layers in the main networks are generated by a shared hypernetwork and embeddings of different layers, which is different from the direct optimized main networks.

\begin{table*}[t]
\centering
\caption{Clean and adversarial accuracy against multiple perturbations of our model with different embedding dimensions of hypernetworks. The red upward arrow ($\textcolor{red}{\uparrow}$) and green downward arrow ($\textcolor{green}{\downarrow}$) denote the increased and decreased  percentage of the parameter number compared with the basic architecture, respectively.}
\vspace{-0.2cm}
\resizebox{\linewidth}{!}{
\begin{tabular}{cccccccccrc}
\toprule[1pt]
Dataset &  Embedding Dim. &  Acc$_\text{clean}$ &  Acc$_{\ell_\infty}$ &  Acc$_{\ell_2}$ &  Acc$_{\ell_1}$ &  Acc$^\text{max}_\text{adv}$ &  Acc$^\text{avg}_\text{adv}$ &  Acc$_\text{trade-off}$ &  \multicolumn{1}{c}{\# of Paras} &  Paras Saving \\
\cline{1-11}
\multirow{4}{*}{\begin{tabular}{c}CIFAR-10\\(ResNet-50)\end{tabular}}     & 32  & 78.93\% & 37.91\% & 58.95\% & 65.37\% & 38.06\% & 54.12\% & 55.56\% & 0.800M    & $\textcolor{green}{\downarrow}96.60\%$\\
                              & 64  & 80.02\% & 39.75\% & 59.92\% & 66.42\% & 39.10\% & 55.67\% & 56.81\% & 1.765M  & $\textcolor{green}{\downarrow}92.51\%$ \\
                              & 128 & 82.28\% & 40.33\% & 60.28\% & 68.13\% & 40.78\% & 56.32\% & 58.02\% & 4.874M                      & $\textcolor{green}{\downarrow}79.30\%$ \\
                              & 256 & 82.34\% & 40.40\% & 60.33\% & 68.17\% & 40.85\% & 56.30\% & 58.06\% & 15.812M & $\textcolor{green}{\downarrow}32.85\%$ \\
\cline{1-11}
\multirow{4}{*}{\begin{tabular}{c}TinyImageNet\\(ResNet-18)\end{tabular}} & 32  & 52.42\% & 27.36\% & 39.71\% & 43.71\% & 27.49\% & 36.93\% & 37.94\% & 0.634M   & $\textcolor{green}{\downarrow}94.38\%$ \\
& 64  & 53.08\% & 27.62\% & 39.84\% & 44.15\% & 27.64\% & 37.19\% & 38.25\% & 1.318M                      & $\textcolor{green}{\downarrow}88.31\%$ \\
& 128 & 53.19\% & 27.67\% & 40.03\% & 44.24\% & 27.74\% & 37.31\% & 38.36\% & 3.866M                      & $\textcolor{green}{\downarrow}65.72\%$ \\
& 256 & 53.24\% & 27.69\% & 40.08\% & 44.26\% & 27.75\% & 37.34\% & 38.39\% & 13.679M                     & $\textcolor{red}{\uparrow}21.31\%$ \\
\bottomrule[1pt]
\end{tabular}
\label{tab.zdim_ablation}
}
\vspace{-0.4cm}
\end{table*}

%------------------ ----------------------------------------------------------------------------
\subsection{Lowest-Entropy Inference}
In the inference stage, we cannot know about which adversarial method the attackers use to perturb the legitimate input. This leaves us with the question: \emph{which output from $\mathcal{F}_\mathcal{H}$ is suitable to serve as the final inference output?} Inspired by information theory \cite{Shannon1948}, one can use the concept of entropy to describe how chaotic a system is: higher entropy means more chaotic and vice versa. Here, we can use entropy to describe how certain an inference is: lower entropy denotes more certain and vice versa. For every inference, we select the output with the lowest entropy as the final model output. This inference strategy can be formulated as (here, we use $f_{\mathcal{H}_p}$ to denote $f_{\mathcal{H}_p(z_p)}$ for simplicity)
\begin{equation}
\small
\mathcal{F}_\mathcal{H}(x_t)=\mathop{\mathrm{argmin}}\limits_{f_{\mathcal{H}_p}(x_t)}-\sum_{k=1}^{|\mathcal{C}|}f^k_{\mathcal{H}_p}(x_t)\log f^k_{\mathcal{H}_p}(x_t),~p\in \mathcal{A},
\end{equation}
where $x_t$ represents the test data; $|\mathcal{C}|$ is the total number of all classes; $f^k_{\mathcal{H}_p}(x_t)$ is the $k$-class probability of the hypernetwork $\mathcal{H}_p$. The schematic of the inference process is shown in the supplementary material.

\section{Experiments}\label{sec:exp}

\begin{figure}[t]
\centering
\subfigure[CIFAR-10]{
    \begin{minipage}[b]{0.477\linewidth}
        \includegraphics[width=0.99\textwidth]{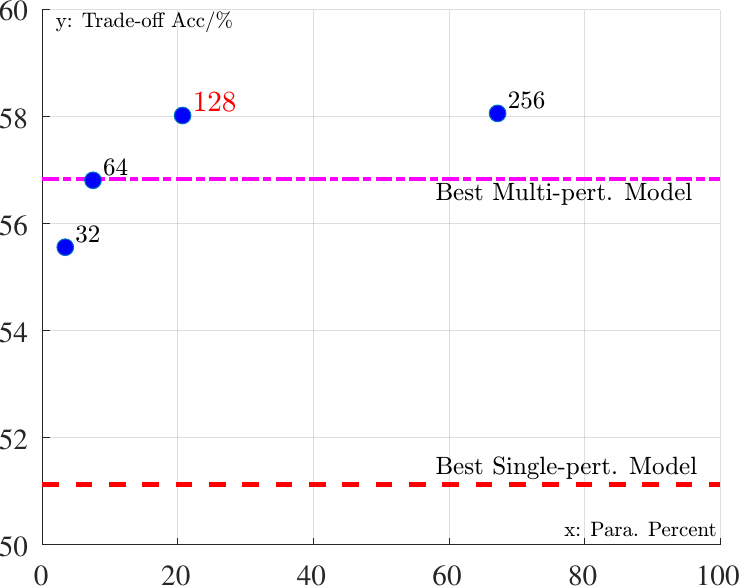}
    \end{minipage}}
\subfigure[TinyImageNet]{
    \begin{minipage}[b]{0.477\linewidth}
        \includegraphics[width=0.99\textwidth]{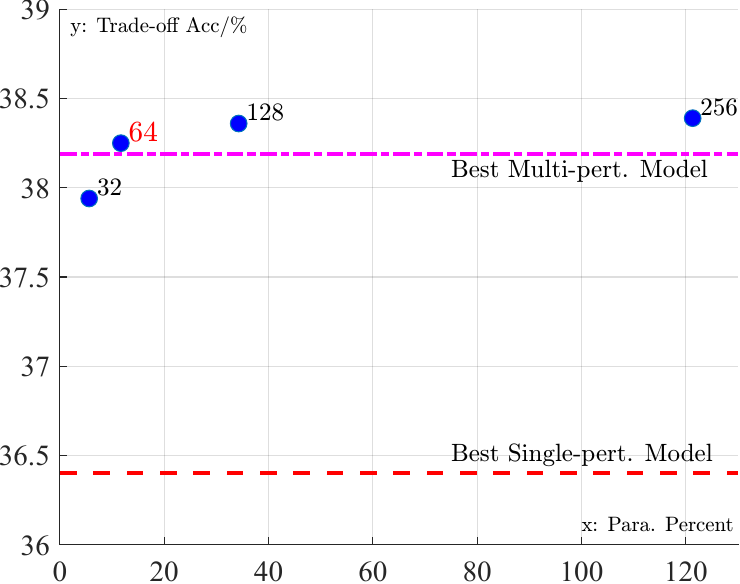}
    \end{minipage}}
\vspace{-0.2cm}
\caption{Trade-off accuracy w.r.t the parameter percentage of the basic network for CIFAR-10 and TinyImageNet datasets. The blue point with red typeface is the utilized embedding dimension in Table \ref{tab.main_results}.
% Here, the x-axis is the parameter percentage of the basic network, while the y-axis is the multi-perturbation robustness trade-off accuracy. We also plotted the best accuracy lines of single-perturbation robust model (red dashed line) and multi-perturbation robust model (magenta dotted dashed line).
}
\label{fig.z_dim_ab}
\vspace{-0.4cm}
\end{figure}

%In this section, we display experimental results on using our proposed PSAT to achieve simultaneous robustness against multiple perturbations (e.g., $\ell_\infty$, $\ell_2$ and $\ell_1$) on three benchmark datasets. In the following, we first compared the proposed method with the previous state-of-the-art methods. Then, we ablated the important design elements of PSAT.

\subsection{Experimental Setup}
\noindent\textbf{Datasets.}\quad
We conducted extensive experiments on the below datasets: CIFAR-10 \cite{Krizhevsky2009}, SVHN \cite{Netzer2011}, TinyImageNet \cite{Russakovsky2015}. CIFAR-10 includes $50,000$ training images and $10,000$ test images with 10 classes. SVHN consists of $73,257$ training images and $26,032$ test images of colored digital numbers in natural scene images with 10 classes. TinyImageNet is the subset of ImageNet \cite{Russakovsky2015} dataset, containing 500 training images, 50 validation images and 50 test images for each class (the total number of classes is 200), respectively. Images on CIFAR-10 and SVHN are sized $32\times32$, and images on TinyImageNet are with size of $64\times64$.

%--------------------------------------------------------------------------------
\noindent\textbf{Baselines and attacks.}\quad
We compared our proposed method with the standard training method (NAT), single-perturbation adversarial training defence methods \cite{Madry2018} for $\ell_\infty$ ($\text{AT}_\infty$), $\ell_1$ ($\text{AT}_1$) and $\ell_2$ ($\text{AT}_2$) norms. Besides, we also considered the state-of-the-art multi-perturbation defence methods: worst-case adversarial training ($\text{AT}_\text{max}$) \cite{Tramer2019} , average adversarial training ($\text{AT}_\text{avg}$) \cite{Tramer2019}, multi steepest descent adversarial training ($\text{AT}_\text{msd}$) \cite{Maini2020} and meta-noise generation method with adversarial consistency ($\text{AT}_\text{mng}$) \cite{Madaan2021}. For better illustrate the integrated robustness performance against multiple perturbations, we used a trade-off evaluation metric, \emph{trade-off accuracy} ($\text{Acc}_\text{trade-off}$), which is defined as the average accuracy of all the adversarial accuracy against multiple perturbations. Due to the space limit, we show more details (such as implementation details and parameter setup) about the experiment setup in the supplementary material.

\begin{table}[t]
\centering
\caption{Clean and adversarial accuracy against multiple perturbations of single hypernetwork model, our model, and our model plus the standard trained hypernetwork model.}
\vspace{-0.2cm}
\resizebox{\linewidth}{!}{
\begin{tabular}{lccccccc}
\toprule[1pt]
Method & Acc$_\text{clean}$ & Acc$_{\ell_\infty}$ & Acc$_{\ell_2}$ & Acc$_{\ell_1}$ & Acc$^\text{max}_\text{adv}$ & Acc$^\text{avg}_\text{adv}$ & Acc$_\text{trade-off}$ \\
\cline{1-8}
$\mathcal{H}_{\mathrm{NAT}}$ & 90.24\% & 0.00\%  & 0.21\%  & 0.08\%  & 0.00\%  & 0.10\%  & 15.10\% \\
$\mathcal{H}_{\ell_\infty}$ & 84.05\% & 42.48\% & 53.34\% & 23.72\% & 23.39\% & 39.85\% & 44.47\% \\
$\mathcal{H}_{\ell_2}$ & 86.85\% & 26.71\% & 62.70\% & 52.11\% & 25.14\% & 47.17\% & 50.11\% \\
$\mathcal{H}_{\ell_1}$ & 89.02\% & 0.02\%  & 0.99\%  & 77.70\% & 0.00\%  & 26.24\% & 32.33\% \\
\cline{1-8}
PSAT & 82.28\% & 40.33\% & 60.28\% & 68.13\% & 40.78\% & 56.32\% & 58.02\% \\
~+ $\mathcal{H}_{\mathrm{NAT}}$ & 90.18\% & 0.00\%  & 0.12\%  & 0.01\%  & 0.00\%  & 0.04\%  & 15.06\% \\
\bottomrule[1pt]
\end{tabular}
\label{tab.member_ablation}
}
\vspace{-0.3cm}
\end{table}

%----------------------------------------------------------------------------------------------------
\subsection{Multi-Perturbation Robustness Comparison} \label{sec.exp.main}

\noindent\textbf{Evaluation on CIFAR-10 dataset.}\quad
The experimental results for CIFAR-10 dataset are shown in Table \ref{tab.main_results}, from which we can see that our proposed method obtains comparable performance compared with the single-perturbation expert model, while for the attack that the single-perturbation expert model was not trained on, our method outperforms the single-perturbation expert model by a great margin. For the multi-perturbation defence, our method achieves the state-of-the-art adversarial accuracy for the average adversarial evaluation. Additionally, for the robustness trade-off, our method significantly outperforms all other counterparts. Furthermore, our method saves about $79.30\%$ of parameters compared with the baseline network. For more experiments, please refer to the supplementary material.

%--------------------------------------------------------------------------------
\noindent\textbf{Evaluation on SVHN dataset.}\quad
Table \ref{tab.main_results} displays the comparative results on SVHN dataset, which showcases that although our proposed method cannot achieve the best robust performance in any attack evaluation, it achieves comparable adversarial robustness very close to the best one so that it can obtain the state-of-the-art robustness trade-off among all attacks. Here, we used the same network architecture as that for CIFAR-10 dataset, the number of model parameters is same as CIFAR-10 dataset: our method saves $79.30\%$ of parameters compared with the basic network architecture.

%--------------------------------------------------------------------------------
\noindent\textbf{Evaluation on TinyImageNet dataset.}\quad
We also verified the performance of our proposed method on complicated dataset, TinyImageNet. The experimental results are shown in Table \ref{tab.main_results}, where we found that our method can still achieve the stat-of-the-art adversarial robustness against multiple perturbations in such big dataset. In addition, our method saves up to $88.31\%$ of parameters compared with the baseline network.

\subsection{Ablation Study}

\noindent\textbf{Embedding dimension.}\quad
To evaluate the influence of the embedding dimension towards robustness and parameter number, we set different embedding dimensions and retrained our models. The results are shown in Table \ref{tab.zdim_ablation} and Figure \ref{fig.z_dim_ab}. It is evident that larger embedding dimension results in better robustness trade-off. However, too large dimension only provides little improvement, e.g., for ResNet-50, the 256 dimension model only improves $0.04\%$ of trade-off compared with the 128 dimension model, while the parameter number increases about 11M compared with hat with the 128 dimension model. In Figure \ref{fig.z_dim_ab}, we intuitively showed the robustness performance of different embedding dimensions, the best single-perturbation robust model and the best multi-perturbation robust model. 
%Here, $\text{AT}_{\ell_2}$/$\text{AT}_{\ell_\infty}$ serve as the best single-perturbation robust models, and $\text{AT}_\text{mng}$/$\text{AT}_\text{mng}$ serve as the best multi-perturbation robust models for CIFAR-10/TinyImageNet datasets, respectively. 

\begin{table}[t]
\centering
\caption{Clean and adversarial accuracy against multiple perturbations of our method with two different inference strategy: average strategy and lowest-entropy strategy.}
\vspace{-0.2cm}
\resizebox{\linewidth}{!}{
\begin{tabular}{cccccccc}
\toprule[1pt]
Inference & Acc$_\text{clean}$     & Acc$_{\ell_\infty}$    & Acc$_{\ell_2}$      & Acc$_{\ell_1}$      & Acc$^\text{max}_\text{adv}$     & Acc$^\text{avg}_\text{adv}$      & Acc$_\text{trade-off}$\\
\midrule[0.5pt]
Entropy       & 82.28\% & 40.33\% & 60.28\% & 68.13\% & 40.78\% & 56.32\% & 58.02\%   \\
Average              & 82.65\% & 42.63\% & 61.82\% & 68.70\% & 14.28\% & 57.69\% & 54.63\%  \\
\bottomrule[1pt]
\end{tabular}
\label{tab.inference_strategy}
}
\vspace{-0.4cm}
\end{table}

\begin{table}[t]
\centering
\caption{Trade-off accuracy and parameter number of lightweight networks and our method.}
\vspace{-0.2cm}
\resizebox{\linewidth}{!}{
\begin{tabular}{ccccccc}
\toprule[1pt]
Network                & MNV1    & MNV3-S  & MNV3-L  & GN      & Ours-64 & Ours-128 \\
\midrule[1pt]
Acc$_\text{trade-off}$ & 52.27\% & 49.53\% & 53.69\% & 51.97\% & 56.81\% & 58.02\%  \\
\midrule[1pt]
\# of Paras            & 3.20M   & 1.53M   & 4.21M   & 3.91M   & 1.76M   & 4.87M   \\
\bottomrule[1pt]
\end{tabular}
\label{tab.lightweight}
}
\vspace{-0.4cm}
\end{table}

%--------------------------------------------------------------------------------
\noindent\textbf{Member analysis.} \quad
We provided the experimental results of standard training and single-perturbation adversarial training with hypernetworks in Table \ref{tab.member_ablation}. The multi-perturbation robustness trade-off is less than satisfactory for single-perturbation adversarial training. Our PSAT aggregated model chose the three single-perturbation adversarial training hypernetworks ($\mathcal{H}_{\ell_\infty}$, $\mathcal{H}_{\ell_2}$ and $\mathcal{H}_{\ell_1}$) as the members, while when appending the standard training hypernetwork into the aggregated model, the clean accuracy increases about 8\%. However, the adversarial accuracy will plummet into zero, probably because the standard trained hypernetwork ($\mathcal{H}_{\mathrm{NAT}}$) will dominate the lowest-entropy output when attacked by adversarial examples.

\begin{table}[t]
\centering
\caption{Trade-off accuracy and parameter number of single models, direct aggregation and our method.}
\vspace{-0.2cm}
\resizebox{\linewidth}{!}{
\begin{tabular}{clcrc}
\toprule[1pt]
Model                         & Method          & Acc$_\text{trade-off}$ & \# of Paras & Paras Saving                           \\
\midrule[0.5pt]
\multirow{4}{*}{Single Model} & AT$_\text{max}$ & 55.48\%                & 23.547M     & -                                      \\
                              & AT$_\text{avg}$ & 54.67\%                & 23.547M     & -                                      \\
                              & AT$_\text{msd}$ & 54.97\%                & 23.547M     & -                                      \\
                              & AT$_\text{mng}$ & 56.84\%                & 23.552M     & $\textcolor{red}{\uparrow}0.02\%$      \\
\midrule[0.5pt]
Direct Aggregate              & -               & 59.74\%                & 70.642M     & $\textcolor{red}{\uparrow}200\%$       \\
\midrule[0.5pt]
\multirow{2}{*}{Ours}         & z-64            & 56.81\%                & 1.765M      & $\textcolor{green}{\downarrow}92.51\%$ \\
                              & z-128           & 58.02\%                & 4.874M      & $\textcolor{green}{\downarrow}79.30\%$ \\
\bottomrule[1pt]
\end{tabular}
\label{tab.direct_aggregate}
}
\vspace{-0.4cm}
\end{table}

\subsection{Further Analysis}

\noindent\textbf{Inference strategy.}\quad
Another popular inference strategy is to use the average output as the final output. We performed experiments on CIFAR-10 dataset to compare the average strategy and the lowest-entropy strategy. The results are shown in Table \ref{tab.inference_strategy}. From the table, we can find that the average inference strategy achieves better performance than the lower-entropy strategy on the clean accuracy, single-perturbation adversarial accuracy and adversarial accuracy against the average manner, while the  adversarial accuracy against the worse-case manner decreases rapidly. Thus, the multi-perturbation robustness trade-off with the lower-entropy strategy is still better than that of the average strategy.

\noindent\textbf{Comparison with lightweight networks.}\quad
Besides, we adversarially trained several multi-perturbation robust lightweight models for comparison. We used quick stochastic adversarial training [similar to \cite{Madaan2021}] on the CIFAR-10 dataset. The selected lightweight models have comparable parameters with our method (Ours-64 and Ours-128): MobileNetV1 (MNV1) \cite{Howard2017}, MobileNetV3-Small as well as MobileNetV3-Large  (MNV3-S and MNV3-L) \cite{Howard2019} and GhostNet (GN) \cite{Han2019}. The results are shown in Table \ref{tab.lightweight}. From the table, we can see that directly simplifying network structure will hamper multi-perturbation robustness trade-off. \cite{Su2018} also has a similar conclusion: \emph{for a similar network architecture, increasing network depth slightly improves robustness}.

\noindent\textbf{Comparison with direct aggregation.}\quad
Due to the theoretical limits to multi-perturbation robustness for a single model [37], a single model can only achieve a sub-optimal multi-perturbation robustness trade-off. Of cause, directly aggregating single-perturbation robust models improves the robustness trade-off, while it also incurs a linear increase in the number of parameters. Here, we conducted experiments on CIFAR-10 to compare the performance of single models, direct aggregation and our method, where the results are displayed in Table \ref{tab.direct_aggregate}. From the table, we can see that the direct aggregation method performs better than single-trained models and our proposed method in trade-off accuracy, while its drawback is also obvious: it greatly increases the number of parameters, which may be unacceptable in some storage-efficient devices/systems.

\section{Conclusion}\label{sec:conclusion}

In this paper, we proposed a new hypernetwork-based model against multiple adversarial perturbations. Most existing robust methods are mainly tailored to defend against single adversarial perturbation, which is impractical in real-world scenarios because attackers may undermine the deployed deep neural networks in any possible adversarial attack. Some works that defend against multiple perturbations are based on a single model, which have their theoretical limit to multi-perturbation robustness. In this paper, we proposed the parameter-saving adversarial training method that utilizes the idea of hypernetworks to generate the parameters of robust deep models. Moreover, we introduced the lower-entropy inference strategy to effectively coordinate the output of member models. Furthermore, extensive experimental results showed that our proposed method achieves better multi-perturbation robustness trade-off with significantly reducing the network parameters.

\bibliography{aaai24}

\end{document}